# 3D Face Recognition using Significant Point based SULD Descriptor

B H Shekar[#1], N Harivinod[#2], M Sharmila Kumari[*3], K Raghurama Holla[#4]

[#]Department of Computer Science, Mangalore University, Mangalore, Karnataka, India.
[1]bhshekar@gmail.com
[2]harivinodn@gmail.com
[4]raghu247@gmail.com

[*]Department of Computer Science & Engineering, P.A. College of Engineering, Mangalore, Karnataka, India.
[3]sharmilabp@gmail.com

*Abstract* - **In this work, we present a new 3D face recognition method based on Speeded-Up Local Descriptor (SULD) of significant points extracted from the range images of faces. The proposed model consists of a method for extracting distinctive invariant features from range images of faces that can be used to perform reliable matching between different poses of range images of faces. For a given 3D face scan, range images are computed and the potential interest points are identified by searching at all scales. Based on the stability of the interest point, significant points are extracted. For each significant point we compute the SULD descriptor which consists of vector made of values from the convolved Haar wavelet responses located on concentric circles centred on the significant point, and where the amount of Gaussian smoothing is proportional to the radii of the circles. Experimental results show that the newly proposed method provides higher recognition rate compared to other existing contemporary models developed for 3D face recognition.**

*Keywords*— **Range Image, SULD descriptor, 3D face recognition**

## I. INTRODUCTION

Automatic recognition of human faces finds numerous applications in surveillance, automatic screening, authentication and human-computer interaction [15]. Face recognition using 2D intensity image is most widespread since it is easy to obtain and operate. However, it is susceptible to the change of pose and illumination. Even though early attempts are focused on 2D face recognition, nowadays the research is mainly focused on 3D model based face recognition. 3D modeling provides pose and illumination invariance representation of faces.

Our work falls into 3D face recognition, because we use 3D or 2.5D face point cloud from a 3D sensor as input and represent them as a range image. Range images are simple representations of 3D information. It is easy to utilize the 3D information of range images because the 3D information of each point is explicit on a regularly spaced grid. Due to these advantages, range images are very promising in face recognition.

The publicly available database [6] provides the 3D data in the form of point clouds. The 3D sensors, used for face capture, produce 2.5D information. In 2.5D data, for a given *(x,y)* coordinate there is only one *z* value is available *i.e.* information of occluded regions does not exist. This data can be easily projected to a 2D image plane and is called *depth image* or *range image*. A range image is a 2D image in which each pixel represents the distance from a point of the face surface to a plane. It encodes the position of the surface directly.

In this work, we address the 3D face recognition problem using range image representation. Range image construction should be preceded with a pose registration module in order to transform faces to a frontal view. We use *Iterative Closest Point (ICP)* algorithm for this purpose. Another key concern in range image construction is the conversion of irregularly sampled 3D points to a regular *(x,y)* grid. To achieve this, interpolation methods are used.

Once the range images are formed, the 3D face recognition problem becomes a 2D image matching problem. Instead of comparing the whole image, we choose significant point comparison to find the match between two range images. The significant points are detected using Hessian based detector. The SULD descriptors for all significant points are computed. The candidate range image is compared with all target range images from the dataset.

The rest of the paper is organised as follows: The survey of 3D face recognition using range image and local point descriptor are given in section II. Section III describes the proposed 3D face recognition model. Database description and experimental results are provided in Section IV and conclusion is presented in section V.

## II. LITERATURE SURVEY

Here we provide a brief survey of range image based 3D face recognition and local descriptor based face recognition.

### A. Range image based 3D face recognition

Recently, we have seen the rapid growth of research in the field of automatic 3D face recognition based on the range images. Achermann et al. [2] extended the eigenface and Hidden Markov Model for 2D face recognition to range images. Heseltine et al. [10] applied the principle component analysis directly to the range images and used the Euclidean distance to measure similarities among the resulting feature vectors. In [11], Hesher et al. presented a procedure for generating range images of faces using data obtained from a 3D scanner and statistically analysed using PCA and ICA. Moreno et al. [19] presented 3D voxel based face representations for face recognition. In order to study the modeling capabilities of depth maps they defined three classes: full, upper-half and left-side facial depth map. Guru and Vikram in [9] proposed 2D pair wise *Fisher linear discriminants (FLD)* as a robust methodology for face recognition.

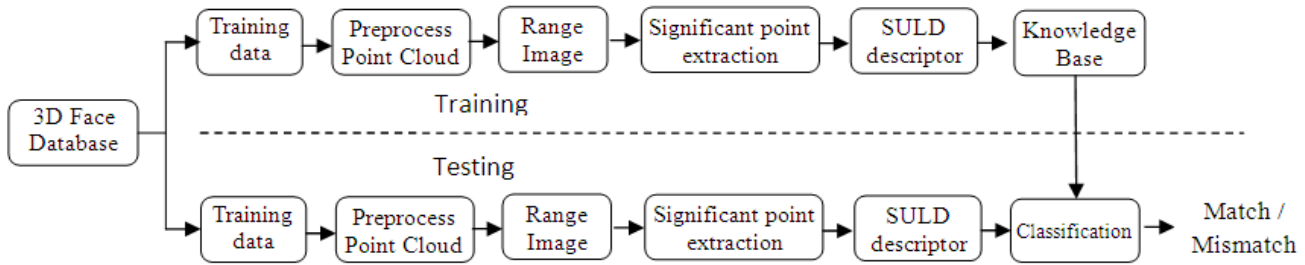

Fig 1. Block diagram of the proposed 3D face recognition system

Thus range images are the widespread way to represent the 3D information and it can store many characteristic features.

### B. Local descriptor based face recognition

Local descriptors are broadly used in the area of computer vision. Ideally these descriptors should be invariant to illumination changes, scaling, rotation, and changes in viewing direction.

Lowe [17] introduced *Scale Invariant Feature Transform (SIFT)* to perform matching between different views of an object. This descriptor is based on the local image gradient, transformed according to the orientation of the significant point to provide orientation invariance. Every feature is a vector of dimension 128, distinctively identifying the neighborhood around the significant point.

SIFT is widely used in face recognition techniques. Mian et al. in [20] used SIFT descriptors for face recognition under illumination and expression variations using 2D and 3D local features. Guo et al. [8] and Lo et al. [16] used 2.5D SIFT descriptor for facial feature extraction in range images. It provides high accuracy and invariance to geometric transformations. The face recognition method proposed by Majumdar and Ward [18] tried to check the number of irrelevant features to be matched thereby reducing the computational complexity in SIFT. Geng and Jiang in [7] used variant of SIFT, called Volume-SIFT (VSIFT) and Partial-Descriptor-SIFT (PDSIFT) for face recognition on 2D faces. Experiments show that the performance of PDSIFT is significantly better than the original SIFT approach. Krizaj et al. in [13] studied adaptation of SIFT features for face recognition under varying illumination. The SIFT descriptors are computed at fixed points on a regular grid and greater robustness to illumination variations is achieved.

In [3], Bay et al. presented *Speeded-Up Robust Features (SURF)*. It is used as detector and descriptor. They used Hessian matrix-based measure for the detector and a distribution-based descriptor for the features in the image. SURF is also used for face recognition. Kim and Dahyot [12] uses SURF for face components detection using support vector machines. Yunqi et al. [24] [25] used SURF for face recognition. They presented 2D face recognition based on FLD to extract the quadratic features on the basis of SURF feature and 3D face recognition by SURF operator based on range Image. In [1], An et al. used SURF for face detection and recognition for human-robot interaction.

Tola et al. [21] [22] proposed DAISY, an efficient dense descriptor for wide baseline stereo matching. Velardo et al. in [23] applied this work to face recognition on 2D images. Zhao et al. in [26] presented SULD descriptor based on the ideas of SURF and DAISY descriptor. It is efficiently computed and used for dense stereo matching.

### III. METHODOLOGY

Here we describe the proposed 3D face recognition model which includes pre-processing, significant point extraction, SULD descriptor computation and descriptor based image matching. The block diagram of the proposed face recognition system is given in Fig. 1.

### A. Pre-processing

The 3D face point clouds are obtained from a 3D face database [6]. Since the face scans of a person differ with pose they need to be registered. We use ICP algorithm [4] to do the registration. The registration process is fully automatic and does not need any manual assistance. In our experiments all the scans of persons are aligned with the first frontal scan of that person.

The face scans usually do not contain corresponding data with respect to the each grid in the range image. So the scattered data is linearly interpolated. We interpolated the data using a Delaunay triangulation. In our implementation, the point nearer to the scanner is identified with highest depth value.

Range image computation is followed by *nose tip* identification. The *nose tip* is identified as the nearest point to the 3D scanner and accordingly it has the maximum range value in the range image. In practice, the *maximum* value of the range images is not just restricted to single point. Instead it contains a region of pixels. In this case the centroid of that region is considered as the desired *nose tip*.

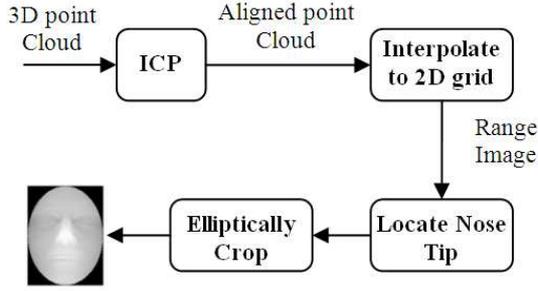

Fig 2. Stages in pre-processing of the point cloud

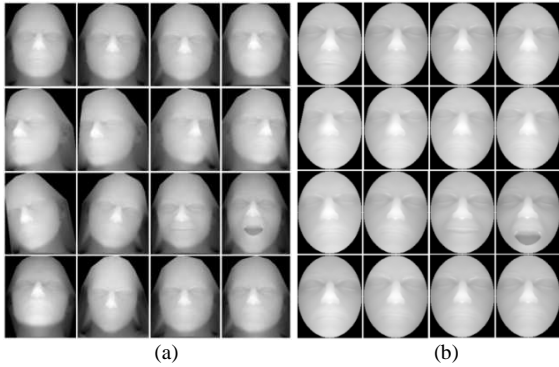

(a)          (b)

Fig 3. All ranges images of a person (a) Before and (b) After pre-processing of the point cloud

Keeping *nose tip* as the centre, the range image is cropped elliptically. The whole process is shown in the Fig 2. Fig 3.a and 3.b show the range images before and after pre-processing respectively.

### B. Descriptor computation and matching

The SULD descriptor associated with each significant point extracted from pre-processed range image is used for recognition purpose. The integral images are used for significant point extraction to reduce the computational burden.

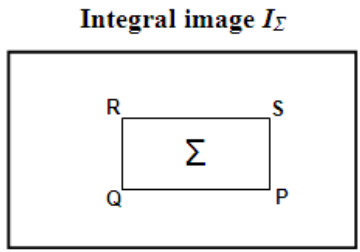

Fig 4. Integral Image *I* with rectangular region *PQRS*. *Σ* shows the summation of the corresponding region *PQRS* from the original image.

*1) Integral Images:*

An integral image is used to compute the sum of intensity values over a rectangular region of the image. Using integral images box type convolution filters are computed quickly. For an image *I*, the integral image $I_\Sigma$ is defined as follows.

$$I_\Sigma(x, y) = \sum_{i=1}^{i \leq x} \sum_{j=1}^{j \leq y} I(i, j) \quad (1)$$

where *x* and *y* represent the row and column number of a pixel in image *I*. $I_\Sigma(x,y)$ is the sum of all *I(x,y)* terms to the left and above the pixel *(x,y)*.

Once the integral image has been computed, it takes just three arithmetic operations to calculate the sum of the intensities over any rectangular area. The sum of rectangular region *PQRS* in image *I* (see Fig. 4.) can be calculated as,

$$\Sigma = I_P + I_R - I_Q - I_S \quad (2)$$

Hence, the calculation time is independent of its size. We use integral images for convolution using box filters.

*2) Significant point extraction*

The significant points are extracted using the SURF detector [3] which is based on the determinant of the Hessian matrix. It uses the determinant of the Hessian given by Lindeberg [14]. For a point *p=(x,y)* in image *I*, the Hessian matrix *H(p,σ)* is the matrix of partial derivatives of the image *I*, in the following format.

$$H(p,\sigma) = \begin{bmatrix} L_{xx}(p,\sigma) & L_{xy}(p,\sigma) \\ L_{xy}(p,\sigma) & L_{yy}(p,\sigma) \end{bmatrix} \quad (3)$$

where $L_{xx}(p,\sigma)$ is the convolution of the Gaussian second order derivative with the image *I* in point *p*, and similarly for $L_{xy}(p,\sigma)$ and $L_{yy}(p,\sigma)$.

Gaussians are widely used for scale-space analysis [3]. Its discrete formations are used in actual implementation. The first two diagrams in Fig. 5 shows discrete Gaussian second order derivative.

The Hessian matrix is computed using the box filters. These approximate second order gaussian derivatives and can be evaluated at a very low computational cost using integral images. The last two diagrams in Fig. 5 illustrate the same. The *9x9* box filters in Fig. 5 are approximations of a Gaussian with *σ=1.2* and represent the lowest scale for computing the blob response maps. Suppose $D_{xx}$, $D_{yy}$ and $D_{xy}$ are the approximations to $L_{xx}$, $L_{yy}$ and $L_{xy}$, the determinant of $H_{approx}$ is

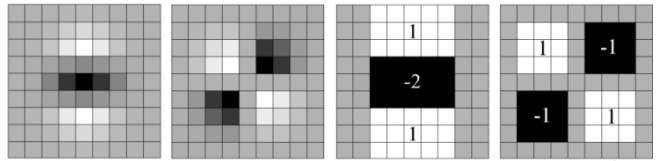

Fig 5. The first two images represent the discretized second order partial derivative in *y* and *xy* direction denoted by $L_{yy}$ and $L_{xy}$ respectively. The approximation for the second order Gaussian partial derivative in *y*- ($D_{yy}$) and *xy*-direction ($D_{xy}$) are shown in last two images. (Courtesy [3])

computed as,

$$\det(H_{approx}) = D_{xx}D_{yy} - (wD_{xy})^2 \quad (4)$$

where $w$ is the weight of the filter responses used to balance the expression for the Hessian's determinant. The approximated determinant of the Hessian represents the response in the image at location $p$.

Lowe [17] implemented the scale spaces using the image pyramid. The images are repeatedly smoothed with gaussian and then sub-sampled in order to achieve a higher level of the image pyramid. But Bay et al. [3], constructed the scale space by up-scaling the filter size rather than iteratively reducing the image size. They used the box filters and integral images. Box filters of any size can be applied on the original image at same speed. We employ the technique proposed in [3].

For a given image, the potential interest points are identified by searching at all scales and based on the stability of the interest points, significant points are extracted. Once the scale space is constructed, the non-maximum suppression in the neighborhood is performed. The maxima of the determinant of the Hessian matrix are then interpolated in scale and image space with the method by Brown et al. [5]. This represents the significant point in the range image.

### 3) SULD descriptor Computation

SULD building process is divided into three stages: computing Haar wavelet response maps, convolving wavelet response maps with Gaussian kernels, and concatenating SULD descriptor by reading the values from convolved response maps.

We compute the Haar wavelet responses for each pixel using $h \times h$ sized filters. Because Haar wavelet filters are box type filters, integral image can be used to reduce computation time. Let $G_x$ be Haar wavelet response in $x$ direction, and $G_y$ be Haar wavelet response in $y$ direction. In order to use the information about the polarity of the intensity changes, the absolute values of the responses *i.e.* $G_{|x|}$ and $G_{|y|}$ are also extracted.

Each response map is convolved several times with Gaussian kernels of different $\Sigma$ values:

$$G_h^{\Sigma} = G_{\Sigma} * G_h \quad (5)$$

where $G_{\Sigma}$ is a gaussian kernel and $G_h^{\Sigma}$ is the convolved response map. Since Gaussian filters are separable, convolutions can be implemented very efficiently. If $G_h^{\Sigma_1}$ has already been computed, then we can efficiently compute $G_h^{\Sigma_2}$ with $\Sigma_2 > \Sigma_1$ by convolving $G_h^{\Sigma_1}$:

$$G_h^{\Sigma_2} = G_{\Sigma_2} * G_h = G_{\Sigma} * G_{\Sigma_1} * G_h = G_{\Sigma^{12}} * G_h^{\Sigma_1} \quad (6)$$

where $\Sigma^{12} = \sqrt{\Sigma_2^2 - \Sigma_1^2}$. Fig. 6 summarizes the required convolutions.

The SULD descriptor for every significant point is computed by picking the values from the convolved response maps. As depicted in Fig. 7, at significant point location, say $(u,v)$, SULD consists of vectors sampled in the neighbourhood around it. These samples located on concentric circles and their amount of gaussian smoothing is proportional to the radius of these circles. Let $h_{\Sigma}(u,v)$ be the vector made up of the values at location $(u, v)$ in the convolved response maps:

$$h_{\Sigma}(u,v) = [G_x^{\Sigma}(u,v), G_y^{\Sigma}(u,v), G_{|x|}^{\Sigma}(u,v), G_{|y|}^{\Sigma}(u,v)]^T \quad (7)$$

where $G_x^{\Sigma}, G_y^{\Sigma}, G_{|x|}^{\Sigma}, G_{|y|}^{\Sigma}$ denote the $\Sigma$ convolved response maps. Before concatenating these vectors to a descriptor, we normalize them to unit vector, and denote the normalized vectors by $\tilde{h}_{\Sigma}(u,v)$. The full descriptor $SULD(u,v)$ for the significant point location $(u,v)$ can be defined as a concatenation of $\tilde{h}_{\Sigma}$ vectors and can be written as:

$$\begin{aligned} SULD(u,v) = \quad & [\tilde{h}_{\Sigma_1}^T(u,v), \\ & \tilde{h}_{\Sigma_1}^T(I_1(u,v,R_1)),...,\tilde{h}_{\Sigma_1}^T(I_N(u,v,R_1)), \\ & \tilde{h}_{\Sigma_2}^T(I_1(u,v,R_2)),...,\tilde{h}_{\Sigma_2}^T(I_N(u,v,R_2)), \\ & \tilde{h}_{\Sigma_3}^T(I_1(u,v,R_3)),...,\tilde{h}_{\Sigma_3}^T(I_N(u,v,R_3))]^T \end{aligned} \quad (8)$$

where $I_j(u,v,R)$ is the location with distance $R$ from $(u,v)$ in the direction given by $j$ when the directions are quantized into N values. Fig. 7 shows the sample locations when N = 8, and SULD descriptor is made up of values extracted from 25 locations and 4 response maps. Therefore, descriptor length is 100 (*i.e.* $4 \times 25$).

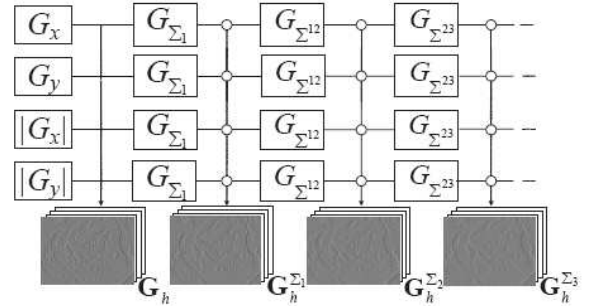

Fig. 6. Convolving response maps representing Eqn. 6. (Courtesy [26])

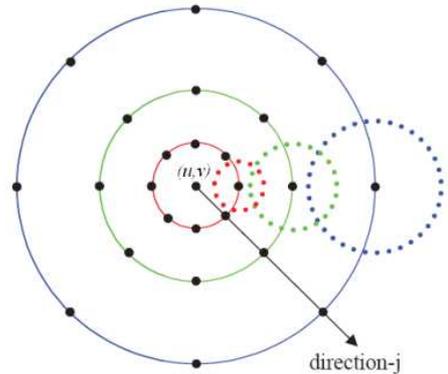

Fig 7. Dark black points stand for the sampling locations for SULD descriptor. The radius of dotted circles is relative to standard deviations of the Gaussian kernels (Courtesy [26]).

TABLE 1.

RECOGNITION ACCURACY OF THE $(2D)^2$ FLD, PCA AND PROPOSED MODEL.

| Test configuration | Total subjects | Training samples | Testing samples | Total samples tested | Rank-1 recognition rate | | |
|---|---|---|---|---|---|---|---|
| | | | | | $(2D)^2$ FLD | PCA | Proposed model |
| T1 | 90 | 1,2,3 | 4 | 90 | 96.67 | 95.56 | 98.89 |
| T2 | 90 | 1,2,3,4 | 11 | 90 | 85.56 | 88.89 | 82.22 |
| T3 | 90 | 1,2,3,4 | 12 | 90 | 52.22 | 46.67 | 76.67 |
| T4 | 90 | 1,2,3,4 | 15,16 | 180 | 94.44 | 97.22 | 100.00 |
| T5 | 90 | 1,2,3,4 | 7,8 | 180 | 94.44 | 87.22 | 94.44 |

*4) Similarity between two face images*

Once we have represented all face images as a set of significant points and their corresponding descriptions, the next step to be carried out is to find similarity measure between two face images. The total number of matches is taken as similarity measure. The matching of two images is done by matching the descriptors of significant points of one image with another. We follow the descriptor matching procedure given by Lowe in [17]. For each significant point of the first image, the best and second best matching points from the second image must be found. If the first match is much better than the second one, the points are said to be alike. Eq. 9 shows how to apply such condition, where points B and C in range image $I_2$ are the best and second best matches, respectively, for point A in range image $I_1$.

$$\frac{\left|D_{I_1}^A - D_{I_2}^B\right|}{\left|D_{I_1}^A - D_{I_2}^C\right|} < threshold \quad (9)$$

In the recognition process we compute the similarity of the test face with all training faces. The training face image corresponding to the highest similarity measure is said to be recognized.

## IV. EXPERIMENTAL RESULTS

### A. Database

Experiments have been conducted on the FRAV3D face database [6]. This database contains the 3D point clouds of 106 persons with 16 captures per person. This includes facial scans with frontal(1,2,3,4), 25° right turn in Y direction (5,6), 5° left turn in Y direction(7,8), severe right turn in Z direction(9), small right turn in Z direction(10), smiling gesture(11), open mouth gesture(12), looking up turn in X direction(13), looking down turn in X direction(14), frontal images with uncontrolled illumination(15,16). The 2D image of all 16 scan of a subject is shown in Fig 2a, where numbering (1-16) starts horizontally from top-left to bottom-right. In our experiments, we have taken the subset of this face database with 90 persons whose all face scans are precisely registered with the first frontal scan.

### B. Results

Experiments have been conducted with different test configurations. $(2D)^2$ FLD and conventional PCA on range images are considered for comparative study. It is observed that the proposed method outperforms the other two methods. In our experiments, the average number of significant points per face range image is 24. Table I shows the results of various combinations of training and testing samples. In T1 naturally we can expect the high recognition rate because all data contains the frontal scan. In T2 and T3, test input contains a gesture. In T4, it is observed that illumination variation does not affect the 3D face recognition. Compared to T1, T5 results have less recognition rate due to self occlusion.

Experiments are also conducted using the leave-one-out strategy taking all 16 face scans of subjects. The results are given in Table II. It is observed that the test face with open mouth mismatches most of the times.

TABLE II.

RECOGNITION ACCURACY USING LEAVE-ONE-OUT STRATEGY.

| Total Persons | Samples tested | Rank-1 recognition rate | | |
|---|---|---|---|---|
| | | $(2D)^2$ FLD | PCA | Proposed model |
| 10 | 160 | 91.25 | 93.12 | 95.00 |
| 20 | 320 | 87.81 | 91.56 | 91.75 |
| 30 | 480 | 86.04 | 90.00 | 90.63 |
| 40 | 640 | 84.84 | 87.81 | 89.84 |
| 90 (All) | 1440 | 81.60 | 86.18 | 86.52 |

## V. CONCLUSION

We developed a model for 3D face recognition based on the SULD descriptor of the significant points. Experimental results show that, the proposed model out performs the conventional holistic face recognition techniques. Future work to be carried out includes the comparison of our proposed model against other significant point detectors and descriptors.

ACKNOWLEDGEMENT

The authors would like to thank the support provided by DST-RFBR, Govt. of India vide Ref. No. INT/RFBR/P-48 dated 19.06.2009.